\documentclass[final]{cvpr}

\usepackage{times}
\usepackage{epsfig}
\usepackage{graphicx}
\usepackage{amsmath}
\usepackage{amssymb}
\usepackage{booktabs}
\usepackage[ruled]{algorithm2e} 
\usepackage{float}
\usepackage{array}
\usepackage{widetext}

\usepackage[pagebackref=true,breaklinks=true,colorlinks,bookmarks=false]{hyperref}

\usepackage{geometry}
\usepackage{xr}
\usepackage{hyperref}

\def\cvprPaperID{5698} 
\def\confYear{CVPR 2021}

\begin{document}

\title{\vspace{-10pt}Real-Time Selfie Video Stabilization\vspace{-5pt}}



\author{Jiyang Yu\textsuperscript{1,3}\quad Ravi Ramamoorthi\textsuperscript{1}\quad Keli Cheng\textsuperscript{2}\quad Michel Sarkis\textsuperscript{2}\quad Ning Bi\textsuperscript{2}\\
\textsuperscript{1}University of California, San Diego\quad \textsuperscript{2}Qualcomm Technologies Inc. \quad \textsuperscript{3}JD AI Research, Mountain View\\
{\tt\small jiy173@eng.ucsd.edu \quad ravir@cs.ucsd.edu \quad {\{kelic,msarkis,nbi\}}@qti.qualcomm.com}
}


\twocolumn[{%
\maketitle
\vspace{-35pt}
\begin{figure}[H]
\hsize=\textwidth 
\centering
\includegraphics[width=\textwidth]{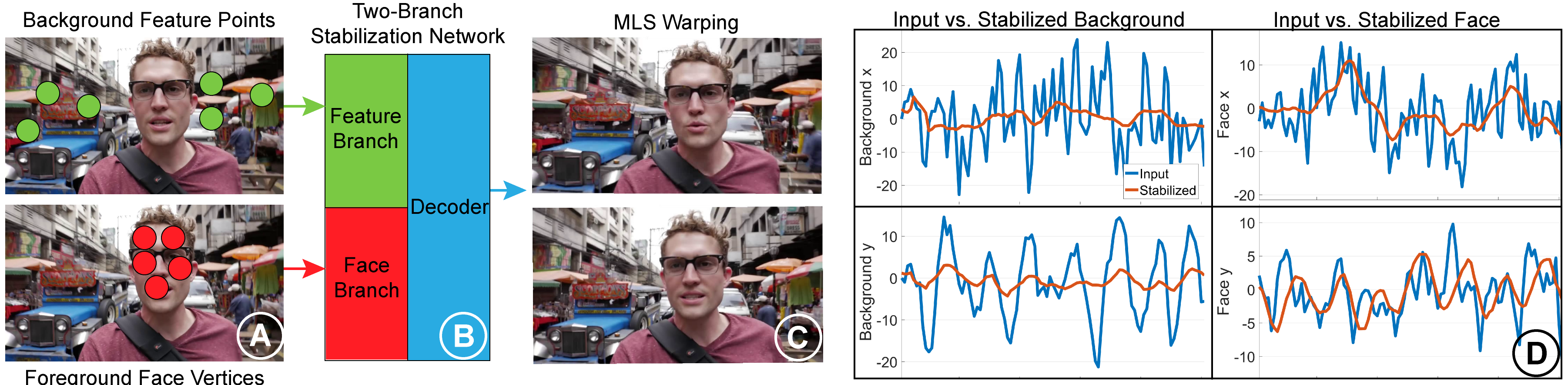}
\caption{\emph{Our method stabilizes selfie videos using \textcircled{A} background feature points and foreground face vertices in each frame. \textcircled{B} The two-branch stabilization network infers \textcircled{C} the moving least squares (MLS) warping for each frame. \textcircled{D} We show the face and background motion of the input vs. our stabilized result. For visualization only, the background tracks are computed from the translation component of the homography between consecutive frames. The face tracks are computed from the centroid of the fitted face vertices in each frame.}}
\label{fig:teaser}
\vspace{-3pt}
\end{figure}
}]

\begin{abstract}
\vspace{-5pt}
We propose a novel real-time selfie video stabilization method. 
Our method is completely automatic and runs at 26 fps. 
We use a 1D linear convolutional network to directly infer the rigid moving least squares warping which implicitly balances between the global rigidity and local flexibility. 
Our network structure is specifically designed to stabilize the background and foreground at the same time, while providing optional control of stabilization focus (relative importance of foreground vs. background) to the users. 
To train our network, we collect a selfie video dataset with 1005 videos\footnote{The dataset is collected at UC San Diego.}, which is significantly larger than previous selfie video datasets. 
We also propose a grid approximation to the rigid moving least squares that enables the real-time frame warping. 
Our method is fully automatic and produces visually and quantitatively better results than previous real-time general video stabilization methods. 
Compared to previous offline selfie video methods, our approach produces comparable quality with a speed improvement of orders of magnitude.
Our code and selfie video dataset is available at \url{https://github.com/jiy173/selfievideostabilization}.
\end{abstract}

\vspace{-10pt}
\vspace{-10pt}
\section{Introduction}\label{sec:intro}
Selfie videos are pervasive in daily communications.
However, capturing high quality selfie video is challenging without specialized stabilization devices like gimbals, which is not convenient, and may not even be allowed in some cases.
On the other hand, from the perspective of algorithms, selfie video stabilization is also challenging.
In general, there are three major steps in video stabilization algorithms.
The first step is to detect the motion in the input video.
Selfie videos have a significant foreground occlusion imposed by human, which is a common limitation of video stabilization algorithms since tracking the frame motion is difficult in the presence of large occlusion.
The second step is to replan/stabilize the motion.
In selfie videos, the motions in foreground/background are usually very different.
Existing selfie video stabilization methods like \cite{steadyface} aim to stabilize the face.
However, stabilizing according to only foreground results in significant shake in the background, and vice versa.
The third step is the warping of the frames.
For selfie videos, the users are sensitive to distortion on the human face.
This requires high rigidity in the foreground warping while maintaining the flexibility in the background warping.

Critically, consumer applications like selfie video stabilization require a significantly fast or even real-time online algorithm to be practical.
This rules out most video stabilization algorithms requiring high overhead pre-processing like SFM~\cite{contentpreserve}, optical flow~\cite{steadyflow,our19,choi} and future motion information~\cite{youtube,subspace}.
A previous selfie video stabilization method~\cite{ourECCV} is an optimization based method and cannot achieve real-time performance.
Although another selfie video stabilization work Steadiface~\cite{steadyface} achieves real-time performance, it only estimates global homography for stabilization and cannot handle non-linear local motions, e.g. rolling shutter.
Additionally, their work also requires gyroscope information.

In this paper, we propose a novel learning based \textit{real-time} selfie video stabilization method.
Our method is fully automatic and requires no preprocessing and user assistance.
The method is designed to tackle the challenges discussed above.
An overview of our method is shown in Fig.~\ref{fig:teaser}.
To achieve real-time performance, our method is purely 2D video stabilization, meaning that our method only depends on the motion of sparse 2D points detected from input video (Fig.~\ref{fig:teaser}\textcircled{A}).
This makes our method significantly faster than the offline selfie video stabilization~\cite{ourECCV}.
In the first step, we avoid the occlusion problem by training a segmentation network to infer the foreground regions and remove the feature points in the foreground.
To take foreground motion into consideration, we use the 3DDFA~\cite{3ddfa} to fit a 3D mesh to video frames.

To warp the original frames into stabilized frames, we use the rigid moving least squares (MLS)~\cite{mls} (Fig.~\ref{fig:teaser}\textcircled{C}).
In our method, we directly use the background feature points as the warp nodes so that the face shape remains undistorted.
Since the original MLS warping is computationally expensive, we use a grid approximation to maintain the real-time performance.
Although faster warping methods exist, e.g. as-similar-as-possible warping in Liu et al.~\cite{bundle}, MLS warping is necessary for our method.
First, traditional grid warping requires an additional hyperparameter to regularize the grid shape.
These terms usually conflict with the motion loss and manually setting the weight between visual distortion and stability is tricky. 
On the other hand, MLS warping guarantees rigidity implicitly and does not require human intervention.
It also preserves the original shape of regions that lack warp nodes.
Second and more importantly, our method is learning based instead of optimization based. 
In the traditional optimization process, it is easy to define the mapping between grid vertices and their
enclosed feature points in the Jacobian. 
However, learning this spatial relation between feature points and grid vertices is difficult and suffers from generalization problems.
In Sec. ~\ref{sec:result} and the supplementary video, we will show that our setup with MLS warping directly defined
on unstructured warp nodes (feature points) is more effective than directly learning the grid like Wang et al.~\cite{wang}.

The core of our method is the stabilization network (Fig.~\ref{fig:teaser}\textcircled{B}).
The network generates the displacement of the warp nodes from the input face vertices and feature points, so that motions of both the foreground (represented by face vertices) and the background (represented by feature points) are minimized.
We also design the network structure so that the user can optionally control the degree of stabilization of the foreground and background on the fly.
In addition, we find that removing activation layers used in traditional neural networks yields better results(supplementary Table~\ref{tab:design}).
The reason is that our formulation requires a linear relation between the input feature point scale and output warp node displacement scale.
Although our network ultimately represents a linear relationship between input feature points and the displacement of output warp nodes, we will show in the supplementary material that direct optimization for this linear relationship is prohibitive in terms of computational efficiency and accuracy (supplementary Table~\ref{tab:optim})\footnote{Note that the objective function we use is non-linear, so a non-linear optimizer needs to be used in any case, rather than simple linear least squares solvers.}.
Training a linear network instead makes the problem tractable, which is similar to how optimizing over non-linear network weights has regularized optimization problems in video stabilization~\cite{our19} and other domains~\cite{prior} in previous works.

The contribution of our paper includes:

1) A novel selfie video stabilization network that enables real-time selfie video stabilization.
Our network directly infers the moving least squares warp from the 2D feature points, stabilizing both the foreground face and background feature motion (Sec.~\ref{subsec:pipeline1} and Sec.~\ref{subsec:pipeline2}).
In Sec.~\ref{subsec:stabilization} we will show that the structure of our network allows an optional control of stabilization focus.

2) Grid approximated moving least squares warping that works at a real-time rate.
For our method, the MLS algorithm with hundreds of warp nodes requires a significant amount of time to warp a frame.
We use a sparse grid to approximate the MLS warping (Sec.~\ref{sec:warp}) that improves the warping speed by two orders of magnitude.
Our entire pipeline is able to stabilize the video at 26fps.

3) A novel large selfie video dataset with per-frame labeled foreground masks.
We will discuss the details of our dataset in Sec.~\ref{subsec:dataset}.
The dataset enables the training of the foreground detection network and the stabilization network in our paper.
We will make our dataset publicly available for face and video related researches.
\section{Previous Work}\label{sec:related}
While video stabilization has been extensively studied, most of the works belong to the offline video stabilization category.
The major reason is that most video stabilization methods rely on temporally global motion information to compute the warping for the current frame.
Recent works using global motion information include the $L_{1}$ optimal camera paths~\cite{youtube}, bundled camera paths~\cite{bundle}, subspace video stabilization~\cite{subspace},  video stabilization using epipolar geometry~\cite{epipolar}, content-preserving warps~\cite{contentpreserve} and spatially and temporally optimized video stabilization~\cite{yswang}.
These works all involve the detection of feature tracks and smoothing under certain constraints.
Some works use optical flow~\cite{steadyflow,our19} or video coding~\cite{codingflow} instead of feature tracking as the motion detection method.
However, they still inherently require future motion information for the global motion optimization.

One may argue that these global optimization based video stabilization methods can be easily modified to online methods by applying a sliding window scheme.
However, note that methods like bundled camera paths~\cite{bundle} only smooth tracks formed by feature points. 
Falsely detected features can easily affect the optimization, especially when the window size is small.
Moreover, \cite{bundle} requires global motion information to achieve the reported result.
One can expect performance to decrease if a short sliding window is applied. 
In Sec.~\ref{sec:result} we will show that \cite{bundle} already generates inferior results than ours using the entire video (Fig.~\ref{fig:visual} and Fig.~\ref{fig:quant}). 
As we will discuss in Sec.~\ref{sec:network}, our pipeline considers all feature points in a window as a whole; the feature points are not only temporally related but also spatially related.
Note that this makes the objective function non-linear, thus we cannot simply use the least squares optimization of \cite{bundle}.
Moreover, our network contains several downsample layers, which effectively blend feature points.
This makes our network robust to individual erroneous features, and it generates satisfactory results with a short 5-frame sliding window. 
%

Deep learning has also been applied to video stabilization in some works.
These attempts include using adversarial networks to generate stabilized video directly from unstabilized input frames~\cite{xu} and estimate a warp grid from input frames~\cite{wang}.
These methods are difficult to generalize to videos in the wild.
Other learning based works (e.g., \cite{choi}) iteratively interpolate frames at intermediate positions.
These works still require optical flow and are prone to artifacts at moving object boundaries.

Some works are more related to the selfie video stabilization context.
An existing selfie video stabilization method~\cite{ourECCV} uses the face centroid to represent the foreground motions while stabilizing the background motions.
However, their method uses the optical flow to detect the background motion and the foreground mask, which is computationally expensive for real-time applications.
Their method is also based on global motion optimization, which makes it impractical in online video stabilization.
Our method does not require the dense optical flow computation and does not require future motion information, therefore is more efficient than their method.

Steadiface~\cite{steadyface} is an online real-time selfie video stabilization method.
They used facial key points as the reference and the gyroscope information as auxiliary to stabilize human faces.
However, their approach uses simple full-frame transformation to warp the frame, which cannot compensate for non-linear distortion like rolling shutter.
Our method uses grid-based MLS warping which provides flexibility to handle non-linear distortions.
Our method also models the face motion more accurately using a face mesh instead of face landmarks in~\cite{steadyface}. 
Due to these limitations, Steadiface~\cite{steadyface} will not produce results comparable with ours by simply adding a hyperparameter to control foreground and background stabilization like our method.
We will show that the quality of our results is significantly better than Steadiface~\cite{steadyface} in Fig~\ref{fig:other}(b) and the supplementary video.

MeshFlow~\cite{meshflow} is an online real-time general video stabilization method.
They use a sparse grid and feature points to estimate the dense optical flow.
However, as a general video stabilization method, they do not consider the foreground/background motion and the large occlusion imposed by the face and body.
This reduces the robustness in the context of selfie videos.

In Sec.~\ref{sec:result}, we will compare our result with selfie video stabilization~\cite{ourECCV}, Steadiface~\cite{steadyface}, MeshFlow~\cite{meshflow} and the state-of-the-art learning based approaches~\cite{choi,wang}.
We also compare with the bundled camera path video stabilization~\cite{bundle} representing a typical offline general video stabilization method as the reference.

\begin{figure}[tbp]
\centering\includegraphics[width=0.48\textwidth]{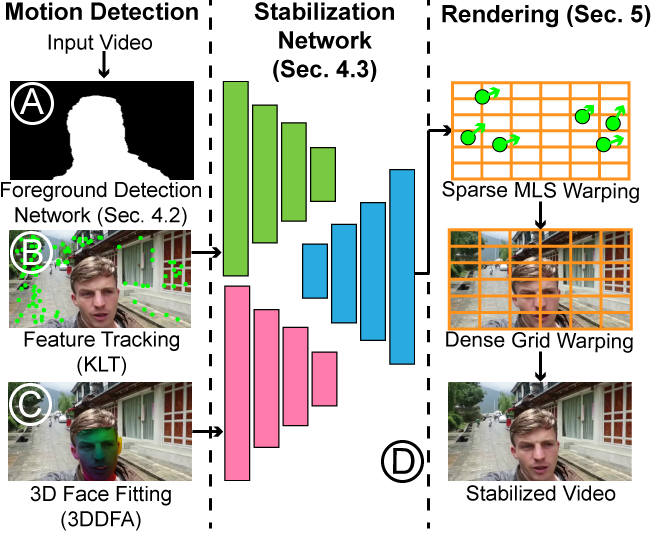}
\caption{\emph{The pipeline of our method. \textcircled{A} We first detect the foreground regions of the input video frame. \textcircled{B} The background motion is tracked using feature points. \textcircled{C} The foreground motion is tracked using 3D face vertices. \textcircled{D} We train a stabilization network to infer the displacement of the MLS warp nodes. Finally, we use a grid to approximate the MLS warping and generate the stabilized frame.}}
\label{fig:pipeline}
\vspace{-15pt}
\end{figure}

\section{Overview of algorithm pipeline}\label{sec:pipeline}
Our pipeline is shown in Fig.~\ref{fig:pipeline}.
The pipeline consists of three major parts: motion detection, stabilization and warping.
In this section, we will introduce these parts separately and provide an overview of the selfie video stabilization process.
For completeness, we summarize the notations used in our paper and supplementary material in supplementary Table~\ref{tab:not}.
The training of the neural networks mentioned below will be discussed in Sec.~\ref{sec:network}.

\subsection{Motion Detection}\label{subsec:pipeline1}
As discussed in Sec.~\ref{sec:intro}, for selfie videos, we seek to stabilize the foreground and background at the same time.
Therefore, both the motion of the face and the background need to be detected.
To distinguish the foreground and the background, we first use a pre-trained foreground detection network to infer a foreground mask $\mathbf{M}_t$ where $\mathbf{M}_t=1$ represents the foreground region of frame $t$.
We show a sample foreground mask in Fig.~\ref{fig:pipeline}\textcircled{A}.
The details regarding the foreground detection network will be discussed in Sec.~\ref{subsec:foreground}.
For the background region where $\mathbf{M}_t=0$, we use the Shi-Tomasi corner detector\cite{KLT} to detect feature points in a frame and the KLT tracker to find their correspondences in the next frame, as shown in Fig.~\ref{fig:pipeline}\textcircled{B}.
We uniformly sample 512 feature points for each frame, since fewer feature points cannot provide enough coverage of frame regions and more feature points will make the pipeline less efficient without significant improvement in warping quality.
We will visually compare the different number of feature point selections in Sec.~\ref{sec:result}.
We denote the selected feature points in frame $t$ as $\mathbf{P}_t \in \mathbb{R}^{2 \times 512}$.
Their correspondences in frame $t+1$ are denoted as $\mathbf{Q}_{t+1} \in \mathbb{R}^{2 \times 512}$.

To detect the motion of the foreground, we fit a 3D face mesh to each frame using 3DDFA proposed in \cite{3ddfa}.
An example of a fitted 3D face mesh is shown in Fig.~\ref{fig:pipeline}\textcircled{C}.
As in the background, we uniformly sample 512 face vertices to represent the face position in a frame.
Furthermore, we only consider the 2D projection of the face mesh in our method.
In this paper, we denote the selected face vertices as $\mathbf{F}_t \in \mathbb{R}^{2 \times 512}$, where $t$ represents the frame index.

\begin{figure}[tbp]
\centering\includegraphics[width=0.48\textwidth]{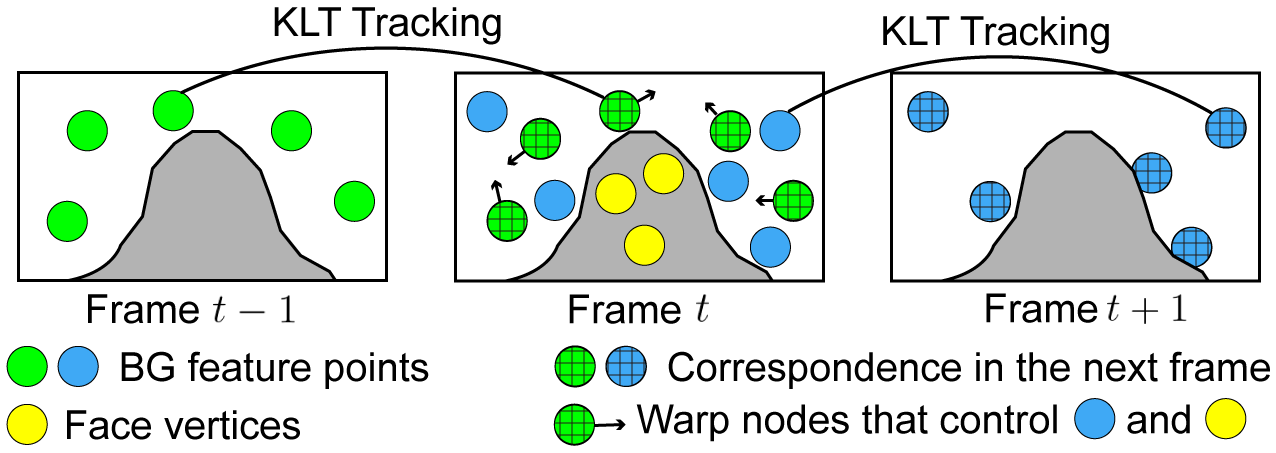}
\caption{\emph{The warping strategy of our method. The background feature points in the same color are in correspondence. The feature points with grid patterns are the warp nodes. The arrows represent the MLS warping operation. During the stabilization, both the feature points $\mathbf{P}_t$(solid blue points) and the face vertices $\mathbf{F}_t$(solid yellow points) are warped by the warp nodes $\mathbf{Q}_t$(grid green points).}}
\label{fig:sequence}
\vspace{-15pt}
\end{figure}

\begin{figure*}[tbp]
\centering\includegraphics[width=\textwidth]{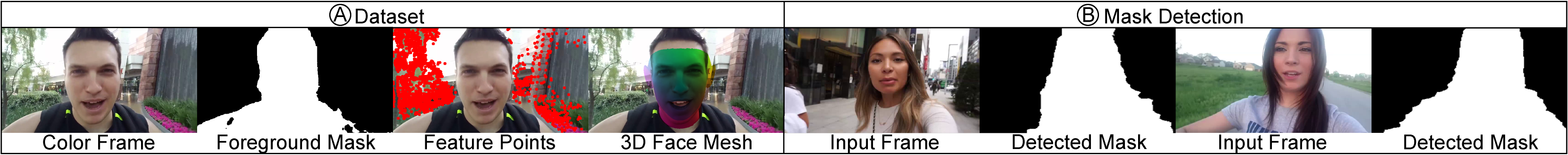}
\caption{\emph{\textcircled{A} Our selfie video dataset. From left to right: color frame, ground truth foreground mask, background feature points, 3D face mesh. \textcircled{B} Examples of the foreground mask detected with our trained foreground detection network.}}
\label{fig:dataset}
\vspace{-13pt}
\end{figure*}

\begin{figure}[tbp]
\centering\includegraphics[width=0.48\textwidth]{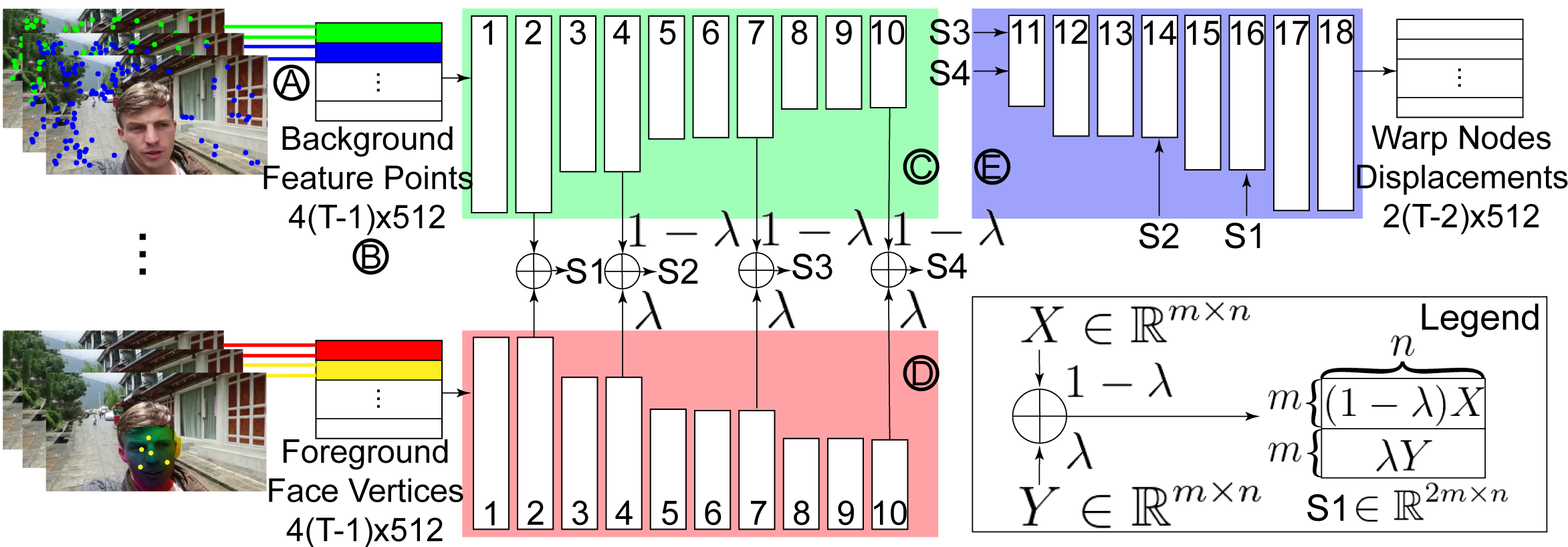}
\caption{\emph{Our stabilization network structure. On the left we show a sequence of input frames. \textcircled{A} The feature points and their correspondences in the next frame are concatenated as a $4\times 512$ tensor. \textcircled{B} The tensors in the same window are concatenated to a large $4(T-1)\times 512$ tensor. The same operation is done for face vertices.  The output of \textcircled{C} the feature branch and \textcircled{D} the face branch of our network are weighted by $\lambda$ and concatenated. \textcircled{E} The decoder outputs the displacements of the warp nodes. The layer parameters are provided in the supplementary material Table~\ref{tab:network}}}
\label{fig:network}
\vspace{-14pt}
\end{figure}

\subsection{Stabilization}\label{subsec:pipeline2}
To stabilize the video, we use the rigid moving least square(MLS) warping\cite{mls} to warp the frames.
In Fig.~\ref{fig:sequence}, we depict the warping strategy of a video sequence.
The moving least square warping requires a set of warp nodes for each frame $t$.
We use the correspondences of detected feature points, i.e., $\mathbf{Q}_{t}$, as the warp nodes for frame $t$ (marked by gridded green dots in Fig.~\ref{fig:sequence}).
Besides all the pixels in frame $t$, the feature points $\mathbf{P}_{t}$ (solid blue dots) and the face vertices $\mathbf{F}_t$ (solid yellow dots) are also warped by $\mathbf{Q}_{t}$ during the stabilization to reflect the change of their positions.

Denote the target location of the warp nodes as $\mathbf{\widehat{Q}}_{t}$, then the rigid MLS warping operation (shown as the arrows in Fig.~\ref{fig:sequence}) can be written as a function $W(\mathbf{v};\mathbf{Q}_{t},\mathbf{\widehat{Q}}_{t})$, where $\mathbf{v}$ is a pixel/feature point/face vertex to be warped.
Denoting each column of a matrix $\mathbf{Q}_{t}$ as $\mathbf{q}_{i,t} \in \mathbb{R}^{2 \times 1}$ where $i\in [1,512]$, the rigid MLS warping procedure is defined by a series of computations.
We included the details of the MLS warping in supplementary material Algorithm~\ref{alg:mls}.
In this paper, we propose a convolutional network (Fig.~\ref{fig:pipeline}\textcircled{D}) to infer the displacements of warp nodes $\mathbf{\widehat{Q}}_{t}-\mathbf{Q}_{t}$.
In Sec.~\ref{subsec:stabilization}, we will discuss the training of this stabilization network.

\subsection{Warping}
More feature points(warp nodes) leads to less warping artifacts but longer time to detect and track.
In our paper, we use 512 feature points as warp nodes in each frame, which is a tradeoff between visual quality and runtime performance.
Details will be discussed in supplementary material Sec.~\ref{sec:warpnode}.
Although the MLS warping can achieve real-time warping with a relatively small number of warp nodes, in our application, warping with hundreds of warp nodes is both time and memory inefficient.
With our implementation of GPU accelerated MLS warping, with 512 warp nodes, a frame of size $448\times832$ must be divided into 16 blocks in order to be fit in a NVIDIA 2080Ti GPU's memory and the warp speed is approximately 1s/frame.
This makes it prohibitive for real-time applications.
To address this issue, we use a grid to approximate the MLS warp field.
This approximation enables real-time performance of our method and yields high-quality visual results.
In Sec.~\ref{sec:warp}, we will demonstrate the details of the grid warping approximation.

\section{Network}\label{sec:network}
In this section, we discuss the details regarding the stabilization network and foreground detection network.
We first present our novel selfie video dataset (Sec.~\ref{subsec:dataset}), then discuss details of the foreground detection network (Sec.~\ref{subsec:foreground}) and stabilization network (Sec.~\ref{subsec:stabilization}).
In the supplementary meterial, we introduce a sliding window scheme to apply our stabilization network to arbitrarily long videos (Sec.~\ref{sec:slide}).

\subsection{Dataset}\label{subsec:dataset}
Although large scale video datasets like Youtube-8M~\cite{youtube8m} have been widely used, public videos with continuous presence of faces are difficult to collect.
We propose a novel selfie video dataset containing 1005 selfie video clips, which is significantly larger than existing selfie video datasets proposed in \cite{ourECCV}(33 videos) and \cite{mobiface}(80 videos).
We first manually collect long vlog videos captured with mobile devices from the Internet.
In these videos, we aim to locate the clips that have stable face presence.
We use the face detector from Dlib~\cite{dlib} to detect faces in each frame, and maintain a global counter to count the number of consecutive frames that contain faces.
If the face can be detected in more than 50 consecutive frames, we cut the raw video into a new clip.
In addition to the regular color videos, our dataset also includes a ground truth foreground mask for each frame.
We manually label the foreground region of the first frame of each video clip, then use Siammask\_E~\cite{siam} to track the foreground object and generate the foreground mask for the video clip.
In addition, we also provide the detected feature points in each frame and their correspondences in the next frame.
Finally, for each frame, we provide the dense 3D face mesh fitted using \cite{3ddfa}.
In Fig.~\ref{fig:dataset}\textcircled{A}, we show a video still, the corresponding foreground mask, the background feature points and the 3D face mesh from our dataset.
Our dataset will be made publicly available upon publication.

\subsection{Foreground Detection Network}\label{subsec:foreground}
Since we have the ground truth mask for our selfie video dataset, training a binary segmentation network is straightforward.
We train an FCN8s network proposed in \cite{fcn} for this segmentation task.
Although there are more advanced structure for segmentation~\cite{LightWeightRF,hardnet}, we find
that FCN8s achieves satisfactory results for our application.
The input of the network is the raw RGB frame, and the output is the binary segmentation mask $M$ mentioned in Sec.~\ref{subsec:pipeline2}.
The training uses Adam optimizer with a $10^{-3}$ learning rate and a binary cross entropy loss.
Figure \ref{fig:dataset}\textcircled{B} provides examples of the inferred masks on video frames outside our dataset.
Note that the inferred mask is not perfect, but it is accurate enough to distinguish the foreground and the background.

\subsection{Stabilization Network}\label{subsec:stabilization}
For a video with $T$ frames, we are able to detect $T-1$ groups of feature points $\mathbf{P}_t$ and their correspondences in the next frame $\mathbf{Q}_{t+1}$ using the KLT tracking mentioned in Sec.~\ref{subsec:pipeline1}.
For each frame, we seek to infer the displacement of warp nodes $\widehat{\mathbf{Q}}_t-\mathbf{Q}_t$ so that the overall motion of the video is minimized.
Formally, the loss function for the background can be written as
\vspace{-5pt}
\begin{equation}\label{eqn:lb}
L_b=\sum_{t=1}^{T-1}\left \| W(\mathbf{P}_t;\mathbf{Q}_t,\mathbf{\widehat{Q}}_t) - \mathbf{\widehat{Q}}_{t+1} \right \|_2
\end{equation}
where $W(\mathbf{P}_t;\mathbf{Q}_{t},\widehat{\mathbf{Q}}_t)$ is the MLS warping function as mentioned in Sec.~\ref{subsec:pipeline2}.
Note that here we apply the MLS warping function to a group of feature points, i.e., each column of $\mathbf{P}_t$ are treated as the coordinates of a pixel and warped by all the warp nodes according to supplementary material Algorithm \ref{alg:mls}.
Since the $\mathbf{P}_t$'s correspondence $\mathbf{Q}_{t+1}$ are the warp nodes for the next frame, so here we should directly use their new position $\mathbf{\widehat{Q}}_{t+1}$.

Similarly, we can also define the foreground loss function using the face vertices:
\begin{small}
\begin{equation}\label{eqn:lf}
L_f=\sum_{t=1}^{T-1}\left \| W (\mathbf{F}_t;\mathbf{Q}_t,\mathbf{\widehat{Q}}_t) - W(\mathbf{F}_{t+1};\mathbf{Q}_{t+1},\mathbf{\widehat{Q}}_{t+1}) \right \|_2
\end{equation}
\end{small}
In this equation, the difference with Eq.~\ref{eqn:lb} is that the face vertices in the next frame $t+1$ are warped by the warp nodes $\mathbf{Q}_{t+1}$.

We also introduce a value $\lambda$ to control the weighting of foreground stabilization and background stabilization.
The complete loss function is defined as:
\begin{equation}\label{eqn:loss}
L=(1-\lambda)L_b+\lambda L_f
\end{equation}
In Eq. (\ref{eqn:loss}), the value $\lambda \in (0,1)$ controls the stabilization focus on foreground versus background.
A larger $\lambda$ means that we tend to stabilize the face more, and a smaller $\lambda$ means we tend to stabilize the background more.
Our method uses $\lambda=0.3$ by default and stabilizes the video automatically.
The user can also change the value online during the stabilization.
In the supplementary video, we will show an example of our network seamlessly changing $\lambda$ during the stabilization.

\noindent \textbf{Network Structure}
Our network structure is inspired by the 2D autoencoder network structure.
However, our formulation only provides sparse feature points as 1D vectors.
The input dimension does not match the 2D network structure.
Moreover, the vanilla autoencoder structure does not provide control over the foreground and background stabilization.
To solve these problems, we design our network as a 1D autoencoder with two input branches.
We demonstrate our network structure in Fig.~\ref{fig:network}.
For simplicity, we will omit the batch dimension in the discussion.
For each frame, the feature points $\mathbf{P}_t\in \mathbb{R}^{2 \times 512}$ and $\mathbf{Q}_t\in \mathbb{R}^{2 \times 512}$ mentioned in Sec.~\ref{subsec:pipeline1} are concatenated in the row dimension, resulting in a frame feature tensor $\mathbf{X}_t \in \mathbb{R}^{4 \times 512}$ as shown in Fig.~\ref{fig:network}\textcircled{A}.
We concatenate the frame feature tensor of $T-1$ frames, forming the feature branch input tensor $\overline{\mathbf{X}} \in \mathbb{R}^{4(T-1) \times 512}$ shown in Fig.~\ref{fig:network}\textcircled{B}.
Similarly, we concatenate the face vertices into the face branch input tensor $\overline{\mathbf{Y}} \in \mathbb{R}^{4(T-1) \times 512}$.
Tensor $\overline{\mathbf{X}}$ and $\overline{\mathbf{Y}}$ are encoded separately with 1D convolutional layers (Figs.~\ref{fig:network}\textcircled{C} and \textcircled{D}), which only convolve with the last dimension of the tensors.
The encoded tensor from different downsample levels are weighted by $\lambda$ and concatenated for skip connection to decoders (Fig~\ref{fig:network}\textcircled{E}), so that the stabilization of foreground and background can be controlled by the user input $\lambda$.
Note that the order of feature points does not affect the network, since we train the network with randomly sampled feature points and face vertices and the encoder downsamples the input and essentially blends the feature points regardless their original order.
The decoder generates the displacements of the warp nodes.
Note that for a length $T$ video, we do not warp the first frame and last frame.
The reason is that the goal of video stabilization is to smooth the original motion, not to eliminate the motion.
Our network is effectively inferring the warp field for the intermediate $T-2$ frames and stabilizes the video instead of aligning all the frames.

\noindent \textbf{Linear Network Design}
Conventional neural networks contain activation layers to introduce non-linearity.
While we started with this design, we found, perhaps surprisingly, that better performance could be obtained by removing the non-linearities(supplementary material Table~\ref{tab:design}).
Specifically, our network does not contain activation layers, which is different from conventional neural networks.
Intuitively, the definition of the loss function(Eq.~\ref{eqn:loss}) requires a linear relationship between the input and the output of the stabilization network, i.e. $N$ times larger feature point coordinates require $N$ times larger output displacement that compensates the motion.

An obvious question to ask here is why training a network is necessary to represent the linear relationship.
In principle, we could pose the problem as an optimization in two alternative ways.
First, it can be modeled as a linear problem in which we solve for a matrix that linearly transforms the vector of input feature points into the output displacement vector.
However, this approach leads to an underdetermined problem with too many variables to be solved for in the full matrix.
Second, we can use a non-linear solver to directly optimize the loss function by solving for the output displacement vector.
However, this solution is prohibitive due to the runtime performance and result quality.

In the supplementary material, we provide a more thorough analysis of our choice of using a linear network.
Briefly, the linear neural network factorizes or regularizes the full matrix optimization (first alternative solution) into smaller
sub-problems that are easier to solve with fewer variables.
Specifically, our analysis includes the necessity of using a network(Sec.~\ref{subsec:linear}), why posing the problem as a non-linear optimization is prohibitive(Sec.~\ref{subsec:optim}) and the performance comparison with traditional neural networks(Sec.~\ref{subsec:nonlinear}).
\newline
\newline
\noindent \textbf{Training}
Our dataset does not contain ground truth stable videos. 
Therefore, our training procedure is unsupervised.
The goal is to learn to minimize the loss function defined in Eq.~\ref{eqn:loss}, i.e. the distances between feature points/face vertices detected in consecutive frames.
Note that the warping is learned solely from groups of unstructured feature points/face vertices. 
To avoid overfitting, we need sufficient diversity in the spatial distribution of these points and motion patterns across the frames. 
Previously discussed efforts we made to satisfy this requirement include a large selfie video dataset(Sec.~\ref{subsec:dataset}) and randomly drawn feature points/face vertices(Sec.~\ref{subsec:stabilization}).
In addition, we further perturb the coordinates of feature points/face vertices using a random affine transformation with rotation between $[-10^{\circ},10^{\circ}]$ and translation between $[-50,50]$ except the first frame and the last frame.
We also generate a random $\lambda$ value between $(0,1)$.
We use Adam optimizer with a $10^{-4}$ learning rate to minimize the loss (Eq.~\ref{eqn:loss}) on length $T$ selfie video clips randomly drawn from our dataset.

\begin{figure}[tbp]
\centering\includegraphics[width=0.48\textwidth]{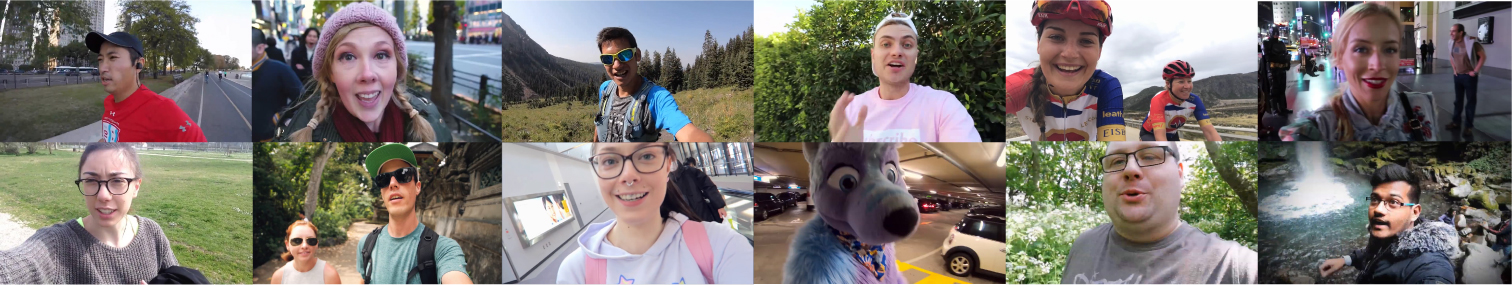}
\caption{\emph{Part of the 25 selfie video examples referred to in Sec.~\ref{sec:result}. Please find complete video stills and corresponding IDs in the supplementary material. Our example videos are selected to cover a variety of challenging scenarios in real applications. }}
\label{fig:example}
\vspace{-10pt}
\end{figure}
\begin{figure*}[tbp]
\centering\includegraphics[width=\textwidth]{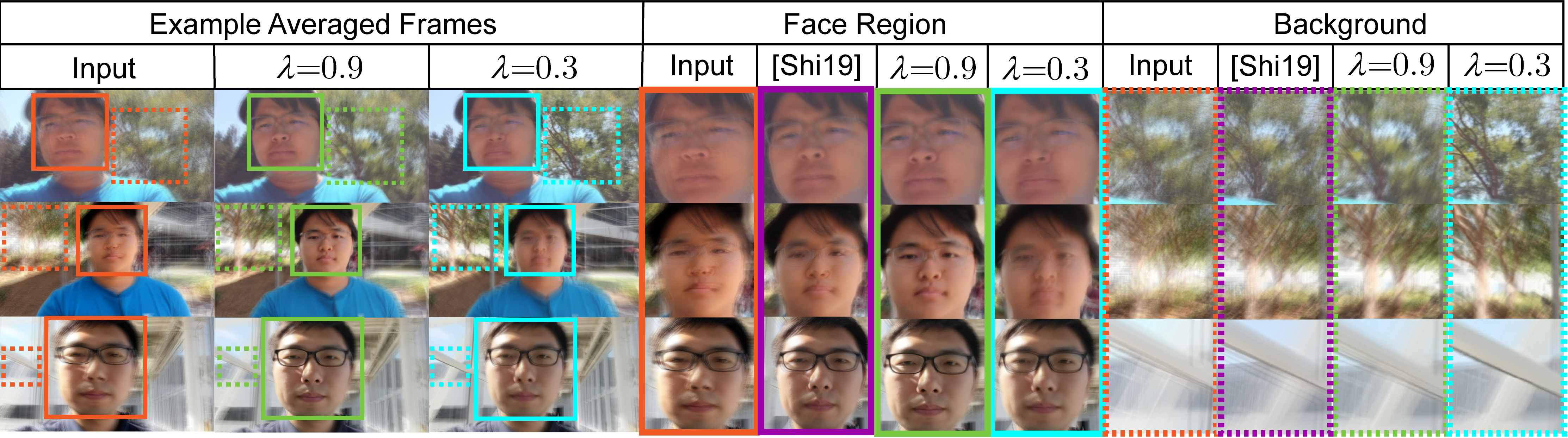}
\caption{\emph{The visual comparison of different values of $\lambda$ in our method and the state-of-the-art real-time face stabilizaiton method Steadiface~\cite{steadyface} using the example videos provided in their work. The images shown are the average of 15 consecutive frames. The face regions and the background regions of the input, the corresponding regions of Steadiface~\cite{steadyface} and our method are shown in the insets on the right.}}
\label{fig:googlevisual}
\vspace{-10pt}
\end{figure*}
\begin{figure*}[tbp]
\centering\includegraphics[width=\textwidth]{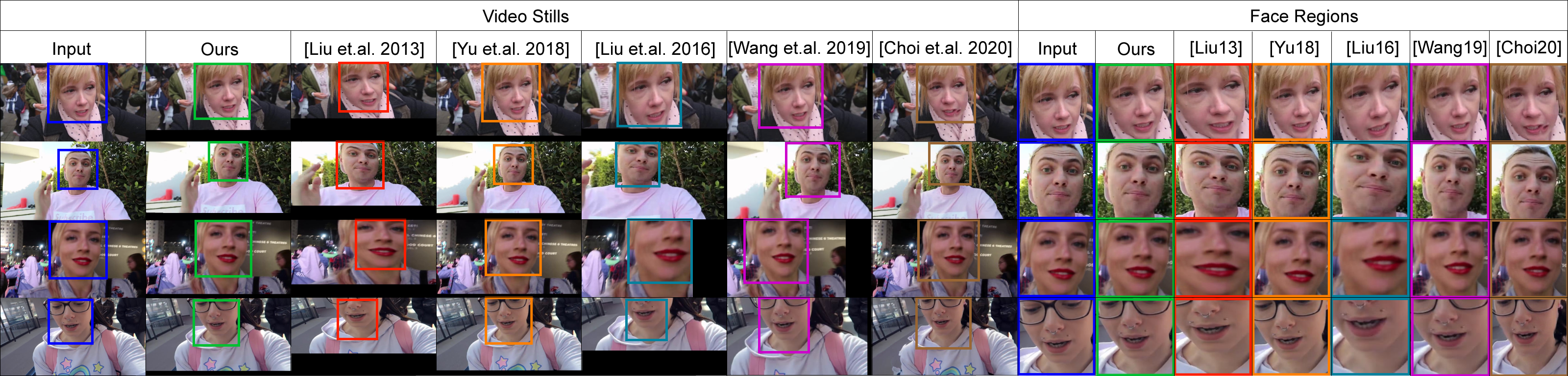}
\caption{\emph{The visual comparison of bundled camera paths~\cite{bundle}, selfie video stabilization~\cite{ourECCV}, MeshFlow~\cite{meshflow}, deep online video stabilization~\cite{wang}, deep iterative frame interpolation~\cite{choi} and our method. The details of the face regions are shown in the insets on the right. We recommend readers to zoom in and observe the details in the images.}}
\label{fig:visual}
\vspace{-15pt}
\end{figure*}

\vspace{-5pt}
\section{Warping Acceleration}\label{sec:warp}
As discussed in Sec.~\ref{sec:pipeline}, using the MLS warping with 512 warp nodes in our case is impractical for real-time application.
To accelerate the warping speed, for the final rendering of the frame, we use a grid to approximate the warp field generated by MLS warping.
Denote a grid vertex in frame $t$ by $\mathbf{g}_j \in \mathbb{R}^{2 \times 1}$, where $j$ is the index of grid vertices.
Each pixel $\mathbf{v}$ can be defined by the bilinear interpolation of the enclosing four grid vertices, denoted by $\mathbf{G} \in \mathbb{R}^{2 \times 4}$: $\mathbf{v}=\mathbf{G}\mathbf{D}$, where $\mathbf{D} \in \mathbb{R}^{4 \times 1}$ is the vector of bilinear weights.

In the first step of rendering, we warp the grid vertices with warp nodes $\mathbf{Q}_t$ and their target coordinates $\mathbf{\widehat{Q}}_t$: $\mathbf{\widehat{g}}_j=W(\mathbf{g}_j;\mathbf{Q}_{t},\mathbf{\widehat{Q}}_{t})$.
Since the grid vertices are sparse, warping with MLS is computationally efficient.
We then densely warp the pixels $\mathbf{v}$ using the MLS warped grid coordinates: $\mathbf{\widehat{v}}=\mathbf{\widehat{G}}\mathbf{D}$, where $\mathbf{\widehat{G}}$ consists of the transformed enclosing four grid vertices $\mathbf{\widehat{g}}_j$.
This step contains only one matrix operation, which can be computed at a real-time rate.
In our experiment, we find the difference between the results generated with the dense MLS warping and grid approximation is negligible.
Our method is not sensitive to the selection of the grid size.
In our experiment, we use a grid size $20\times 20$.
We implemented the grid warping on GPU by parallel sampling the grid with a pixel-wise dense grid, generating a dense warp field.
We then use the dense warp field to sample the video frame, generating the warped frame.
Our implementation of this process takes approximately 4ms/frame, compared to the 1s/frame ground truth dense MLS warping.

\begin{figure}[tbp]
\centering\includegraphics[width=0.48\textwidth]{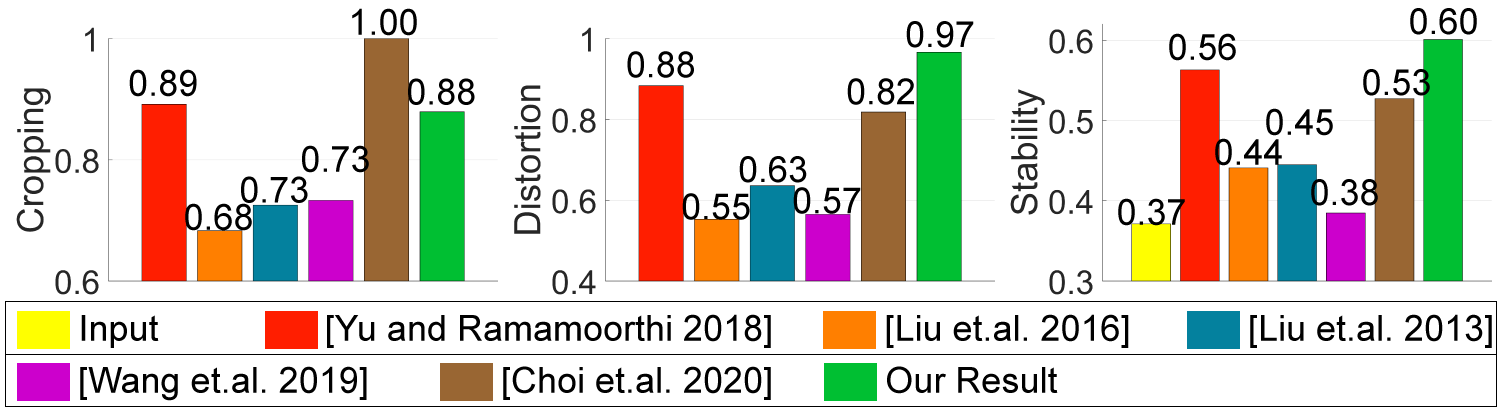}
\caption{\emph{Quantitative comparison of bundled camera paths~\cite{bundle}, selfie video stabilization~\cite{ourECCV}, MeshFlow~\cite{meshflow}, deep online video stabilization~\cite{wang}, deep iterative frame interpolation~\cite{choi} and our method. In these metrics, a larger value indicates a better result. The average values over all the example videos are listed. The complete comparison on individual videos are provided in the supplementary material Fig.~\ref{fig:quant}.}}
\label{fig:quant}
\vspace{-20pt}
\end{figure}
\vspace{-5pt}
\section{Results}\label{sec:result}
In this section, we present the results of our method.
Note that our dataset is cut from a small number of long vlog videos, therefore the faces are from a limited number of people.
Some videos in our dataset also do not actually need to be stabilized (e.g., still camera video).
To show the effectiveness and the ability of generalization of our method, we collect 25 new selfie videos that contain a variety of challenging scenarios in real applications, and are completely separate from our training dataset.
Part of the testing examples are shown in Fig.~\ref{fig:example}.
The complete example video stills with video IDs will be provided in supplementary Fig.~\ref{fig:example}.
The background scenes vary from indoor (example 16, 18, 19), inside of cars (example 7, 12), city (example 1, 2, 8, 9, 10, 13, 15, 21, 22, 23), crowd (example 2, 3, 9, 10, 16, 23, 24) and wild (example 4, 5, 6, 11, 14, 17, 20, 24, 25).
Some of these videos are selected since their content is technically challenging.
These challenges include lack of background features (example 6, 7, 12, 15), dynamic background (example 2, 3, 9, 10, 16, 23, 24), sunglasses (example 4, 7, 14, 15, 21), large foreground occlusion (example 13, 16, 20, 22), face cannot be detected or incomplete face (example 8, 9, 13, 16, 18, 20, 22), multiple faces (example 6, 14) and intense motions (example 1, 23).
Since the dynamics cannot be shown through video stills, we recommend readers to watch our supplementary video.
In the supplementary video, we show the example video clips and our stabilized result side by side. 
As mentioned in Sec.~\ref{sec:related}, we also provide visual and quantitative comparison with the offline selfie video stabilization~\cite{ourECCV}, the real-time selfie video stabilization Steadiface~\cite{steadyface}, the real-time general video stabilization MeshFlow~\cite{meshflow}, the offline general video stabilization bundled camera paths~\cite{bundle} and the state-of-the-art learning-based methods~\cite{choi} and~\cite{wang}.
Since our videos do not contain gyroscope data, we compare with Steadiface~\cite{steadyface} using only the examples provided in their paper.
Apart from the results discussed in this section, we provide more discussion regarding the number of feature points(warp nodes) in supplementary Sec.~\ref{sec:warpnode}, ablation study regarding the FG/BG mask in supplementary Sec.~\ref{sec:ablation} and performance with different input resolution in supplementary Sec.~\ref{sec:framesize}.

\subsection{Value of $\lambda$}
In Fig.~\ref{fig:googlevisual} we show the effect of different values of $\lambda$.
We stabilize the same video clip with $\lambda$ set to 0.3 and 0.9 respectively.
To show the steadiness of the result, we average 15 consecutive frames of the stabilized video.
The less blurry the region is, the more stable it is in the result.
For $\lambda=0.9$, the face regions are less blurry as shown in the green inset, indicating that our network automatically focuses on stabilizing the face.
If we set $\lambda=0.3$, the background regions are less blurry as shown in the cyan inset meaning that the background is more stable.
In our experiment, we use a default value of $\lambda=0.3$, meaning that we stabilize both foreground and background while mainly focusing on the background.

\subsection{Visual Comparison}\label{subsec:visual}
We show sample frames from our examples and the stabilized results in Fig.~\ref{fig:visual}.
Our method stabilizes the frames without introducing visual distortions.
The real-time general video stabilization method~\cite{meshflow} and offline general video stabilization method~\cite{bundle} usually produce artifacts on the face, since they do not distinguish the foreground and the background.
Selfie videos are also challenging for the optical flow estimation in MeshFlow~\cite{meshflow}, since the motion within a mesh cell can be significantly different due to the foreground occlusion.
The learning based method~\cite{wang} generally does not produce local distortions, but tends to generate unstable output video.
Due to the accuracy issue in optical flow and frame interpolation, the other learning based method~\cite{choi} generates artifacts, especially near the occlusion boundaries like face boundaries.
These artifacts are more obvious when observed dynamically in videos.
We recommend the readers to watch the supplementary video for better visual comparison.
We also achieve the same quality visual results as the previous optimization based selfie video stabilization~\cite{ourECCV}.
However, our method is learning-based and runs at the real-time speed, which is orders of magnitude faster compared to their method as we will discuss in Sec.~\ref{sec:speed}.

We also test our method on the examples in Steadiface~\cite{steadyface}, which is the state-of-the-art real-time face stabilization method.
The images shown on the left of Fig.~\ref{fig:googlevisual} are the average consecutive 15 frames of their results.
If we set $\lambda=0.9$ in our method (mainly stabilize the face), we are able to achieve better face alignment.
In addition, we can alternatively set $\lambda=0.3$ in the stabilization network.
The background becomes significantly more stable than the Steadiface~\cite{steadyface} results and our $\lambda=0.9$ results in the averaged frames, indicating that our method is capable of stabilizing the background.
Figure \ref{fig:googlevisual} also indicates that stabilizing the background ($\lambda=0.3$) leads to a slight sacrifice of face stability, since the motion of the foreground and background is different. 
In our supplementary video, we will show that this loss of face stability is visually unnoticeable.

\subsection{Quantitative comparison}\label{subsec:quant}
\begin{figure}[tbp]
\centering\includegraphics[width=0.48\textwidth]{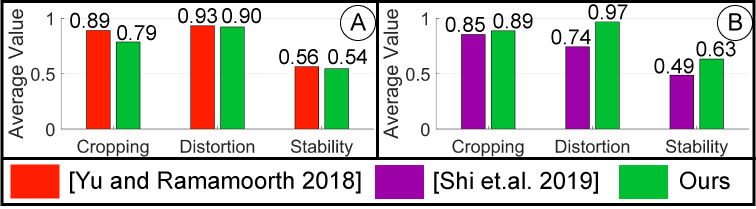}
\caption{\emph{Quantitative comparison with \textcircled{A} selfie video stabilization~\cite{ourECCV} and \textcircled{B} Steadiface~\cite{steadyface} using their datasets respectively. The average values over the entire datasets are plotted. In all the three metrics, a larger value indicates a better result.}}
\label{fig:other}
\vspace{-15pt}
\end{figure}

We use the three quantitative metrics proposed in \cite{bundle} to evaluate the frame size preservation (Cropping), visual distortion (Distortion) and steadiness (Stability) of the stabilization result.
Note that since Steadiface~\cite{steadyface} require gyroscope information to stabilize the video, the quantitative comparison with their method is conducted using their videos and will be discussed in Fig.~\ref{fig:other}\textcircled{B}.

In the left column of Fig.~\ref{fig:quant}, we show the cropping metric comparison.
A larger value represents a larger frame size of the stabilized result.
Although \cite{ourECCV} uses second order derivative objective, their frame size is limited by the motion of the entire video.
Our sliding window only warps the frames with respect to the temporally local motion, so we are still able to achieve similar cropping value while directly using the explicit motion loss in Eq.~(\ref{eqn:loss}).
The frame size of our result is also significantly greater than \cite{meshflow}, \cite{wang} and \cite{bundle}, since the artifacts in their results often cause over-cropping in the final video.
Since \cite{choi} is based on frame interpolation, their cropping score is by default equal to 1.
However, \cite{choi} is essentially an offline method requiring multiple iterations over the entire video. 
In the following discussions, we will show that their distortion and stability score is much worse than ours.

In the middle column of Fig.~\ref{fig:quant}, we show the distortion metric.
This metric measures the anisotropic scaling of the stabilized frame.
A larger value indicates that the visual appearance of the result is more similar to the input video.
Since we warp the frame with grid approximated moving least squares, minimal anisotropic scale was introduced to the result.
The MeshFlow method~\cite{meshflow} and bundled camera paths~\cite{bundle} introduces unexpected local distortion to the frame, which leads to the negative impact on the distortion value.
The learning based methods \cite{wang} and \cite{choi} cannot generalize to selfie videos.
They also produce visual artifacts that lead to even worse distortion values comparing to optimization based methods \cite{meshflow,bundle}.

The right column of Fig.~\ref{fig:quant} shows the stability metric comparison.
A larger stability metric indicates a more stable result.
This is the most important metric for video stabilization.
Comparing with the input (the yellow bar on the left of each example), our method significantly increases the stability in the result.
Our method achieves a comparable result with the optimization based method~\cite{ourECCV} with orders of magnitude improvement in stabilization speed.
We also achieve better stability than \cite{choi,meshflow,bundle,wang}, which is expected since their visual result is not satisfactory as shown in Fig.~\ref{fig:visual}.

To further verify the performance of our method, we also test our method on the selfie videos provided in \cite{ourECCV} and \cite{steadyface}. 
Figure \ref{fig:other} shows the average values of the three metrics above on the selfie video dataset proposed by \textcircled{A}~\cite{ourECCV} and \textcircled{B}~\cite{steadyface}.
Again, our result has a quantitative performance comparable with \cite{ourECCV}.
Our method also performs better than \cite{steadyface} without using the gyroscope information.

\subsection{Stabilization Speed}\label{sec:speed}
Our code is written in Python and runs on a desktop computer with an NVIDIA 2080Ti graphics card.
On average, our method uses 38ms to stabilize a frame of resolution 832x448, which is equivalent to 26fps.
The break down of runtime is 3ms for foreground mask detection, 7ms for the feature detector, 3ms for KLT tracking, 16ms for face mesh detection, 5ms for stabilization network inference, less than 1ms for MLS grid approximation and 4ms for frame warping.
For other video resolutions, we rescale the feature points to match our frame size of 832x448.
The only operation impacted is the grid warping.
However, since the warping is implemented on the GPU, the difference is subtle, e.g. 4ms for HD(1280x720) and 6ms for FHD(1920x1080).
The overall speed is around 40ms/frame for HD and 42ms/frame for FHD.
With frame size 832x448, the average stabilization time of the comparison methods(per frame) are: 4720ms for selfie video stabilization~\cite{ourECCV}, 392ms for bundled camera paths\cite{bundle}, 8ms for Steadiface\cite{steadyface}, 20ms for MeshFlow\cite{meshflow}, 28ms for deep online video stabilization\cite{wang}, 67ms for deep iterative frame interpolation\cite{choi}.
Our method is nearly two orders of magnitude faster than the previous selfie video stabilization~\cite{ourECCV}, and nearly an order of magnitude faster than the traditional optimization based general video stabilization~\cite{bundle}.
Our method is also nearly two times faster than the frame interpolation method~\cite{choi}, since their network involves 2D convolutions.
Also note that ~\cite{choi} is an offline method requiring future frames and multiple iterations through the entire video.

Although our method is slightly slower than MeshFlow~\cite{meshflow} and deep multi-grid warping~\cite{wang}, we have shown in Sec.~\ref{subsec:visual}, Sec.~\ref{subsec:quant} and supplementary video that our method produces significantly better results than theirs.
Our method is also slower than Steadiface~\cite{steadyface}.
However, our method is a purely software video stabilization and requires no gyroscope information, which is not available on some devices, e.g., action cameras.
In addition, since gyroscope information does not provide direct image domain motion, our approach usually yields visually more stable results as we will show in our supplementary video.
As we discussed earlier in Sec.~\ref{sec:related}, our method essentially more accurately models the frame motion than Steadiface~\cite{steadyface}.
Therefore their method does not generate comparable quality as our method.
Also note that our method also runs at a real-time speed without any attempt to optimize the implementation.
We believe that the speed of our pipeline can be further improved by using the GPU memory sharing between feature detection/tracking and neural network operations to avoid repetitive data transferring between CPU and GPU.

\subsection{Limitation}
Our method fails if very few feature points are detected in the background, since our method requires a reasonable number of warp nodes to warp the frame. 
These cases include very dark environments, pure white walls and blue sky.
This is a common limitation for feature tracking based methods~\cite{epipolar,youtube, contentpreserve,subspace,bundle}.
In our method, this can be solved by replacing the feature tracking with the optical flow algorithm with appropriate accuracy and real-time performance.

\section{Conclusions and Future Work}
In this paper, we proposed a real-time learning based selfie video stabilization method that stabilizes the foreground and background at the same time.
Our method uses the face mesh vertices to represent the motion of the foreground and the 2D feature points as the means of background motion detection and the warp nodes of the MLS warping.
We designed a two branch 1D linear convolutional neural network that directly infers the warp nodes displacement from the feature points and face vertices.
We also propose a grid approximation to the dense moving least squares that enables our method to run at a real-time rate.
Our method generates both visually and quantitatively better results than previous real-time general video stabilization methods and comparable results to the previous selfie video stabilization method with a speed improvement of orders of magnitude.

Our work opens up the door to high-quality real-time stabilization of selfie videos on mobile devices.
Moreover, we believe that our selfie video dataset will inspire and provide a platform for a variety of graphics and vision research related to face modeling and video processing.
In the future, we would explore the possibility of learning based selfie video frame completion using our proposed selfie video dataset.

\noindent \textbf{Acknowledgements.} This work was funded by a Qualcomm FMA Fellowship.  We also acknowledge support from the Ronald L. Graham chair and the UC San Diego Center for Visual Computing.

{\small
\bibliographystyle{ieee_fullname}
\bibliography{egbib}
}

\pagebreak


\renewcommand\thetable{\alph{table}}
\renewcommand\thesection{\Alph{section}}
\renewcommand\thefigure{\alph{figure}}



\begin{widetext}
\begin{center}
\textbf{\large Supplementary Material}
\end{center}
\end{widetext}

\setcounter{equation}{0}
\setcounter{figure}{0}
\setcounter{table}{0}
\setcounter{page}{1}
\makeatletter
\renewcommand{\theequation}{S\arabic{equation}}
\renewcommand{\thefigure}{S\arabic{figure}}



\begin{table}%
\caption{Notations in the paper and supplementary material}
\label{tab:not}
\begin{minipage}{\columnwidth}
\begin{center}
\begin{tabular}{ll}
  \toprule
  Symbols          & Explanation\\ \midrule
  $t$              & Frame index\\
  $\mathbf{M}_t$     & Foreground mask\\
  $\mathbf{P}_t$	&  Background feature points\\
  $\mathbf{Q}_{t}$    &  Correspondence of $\mathbf{P}_{t-1}$ in frame $t$\\
  $\mathbf{F}_t$	&  Face vertices\\
  $\mathbf{\widehat{Q}}_{t}$    &  Target coordinate of $\mathbf{Q}_{t}$\\
  $\mathbf{v}$              &  Coordinate of a pixel\\
  $W(\mathbf{v};\mathbf{Q}_{t},\mathbf{\widehat{Q}}_{t})$  &  Rigid MLS warping function\\
  $\mathbf{\widehat{v}}$	&  Warped coordinate of pixel $\mathbf{v}$\\
  $\mathbf{q}_{i}$        &  $i$th column of $\mathbf{Q}_t$\\
  $w_i$						&  MLS weight of $\mathbf{q}_{i,t}$ to pixel $\mathbf{v}$\\
  $\alpha$					&  MLS parameter\\
  $\mathbf{c}$				&  Weighted centroid of $\mathbf{Q}_t$\\
  $\mathbf{\widehat{c}}$				&  Weighted centroid of $\mathbf{\widehat{Q}}_t$\\
  $\mathbf{q}^*_i$           & Vector from $\mathbf{c}$ to $\mathbf{q}_{i,t}$\\
  $\mathbf{\widehat{q}}^*_i$           & Vector from $\mathbf{\widehat{c}}$ to $\mathbf{\widehat{q}}_{i}$\\
  $\mathbf{A}_i$  			& Transformation matrix of $\mathbf{\widehat{q}}^*_i$\\
  $\mathbf{g}_j$ 			& $j$th grid vertex\\
  $\mathbf{G}$				& Grid vertices enclosing $\mathbf{v}$\\
  $\mathbf{D}$				& Bilinear weights of $\mathbf{v}$ with respect to $\mathbf{G}$\\
  \bottomrule
\end{tabular}
\end{center}
\end{minipage}
\vspace{-15pt}
\end{table}%

\begin{algorithm}[t]
\SetAlgoLined
\SetKwInOut{Input}{Input}
\SetKwInOut{Output}{Output}
\Input{Source coordinates of a pixel $\mathbf{v}$, source node coordinates $\mathbf{Q}$ and target node coordinates $\mathbf{\widehat{Q}}$}
\Output{Target coordinates of a pixel $\mathbf{\widehat{v}}$}
\normalsize
\For{$i\gets1$ \KwTo 512}{
    $w_i=1/\left | \mathbf{v}-\mathbf{q}_{i} \right |^{2\alpha}$
    }
    
$\mathbf{c}=\left (\sum _{i=1}^{512}w_i \mathbf{q}_i\right )/\left (\sum _{i=1}^{512}w_i\right )\; \; \; $
$\mathbf{\widehat{c}}=\left (\sum _{i=1}^{512}w_i \mathbf{\widehat{q}}_i\right )/\left (\sum _{i=1}^{512}w_i\right )$

\For{$i\gets1$ \KwTo 512}{
  $\mathbf{q}^*_i=(\mathbf{q}_i-\mathbf{c})^T\; \; \;$
  $\mathbf{\widehat{q}}^*_i=(\mathbf{\widehat{q}}_i-\mathbf{\widehat{c}})^T$ \\
  $\mathbf{A}_i=w_i\binom{\mathbf{q}^*_i}{-{\mathbf{q}^*_i}^{\perp}}\left (\mathbf{v}-\mathbf{c}\; \; \; -(\mathbf{v}-\mathbf{c})^{\perp}\right )$,\\
  where $\perp$  is an operator on 2D vector $(x,y)^{\perp}=(-y,x)$

}
$\mathbf{\widehat{v}}=\left | \mathbf{v}-\mathbf{c} \right |\left (\sum _{i=1}^{512}\mathbf{A}_i \mathbf{\widehat{q}}^*_i \right)/\left | \sum _{i=1}^{512} \mathbf{A}_i  \mathbf{\widehat{q}}^*_i \right |+\mathbf{\widehat{c}}$

\caption{The rigid MLS warping algorithm $W(\mathbf{v};\mathbf{Q},\mathbf{\widehat{Q}})$}
\label{alg:mls}
\end{algorithm}

\section{Implementation Details}\label{sec:detail}
\subsection{MLS warping process}
Table~\ref{tab:not} summarizes the notations used in the main paper and this supplementary material.
Algorithm~\ref{alg:mls} provides the MLS warping process referred in Sec.~\ref{subsec:pipeline2}.
Since the MLS warping is not related to the time dimension, we omit the time subscript $t$ for simplicity.
In Algorithm~\ref{alg:mls}, we use relatively small $\alpha=0.3$ to maintain a smooth warp field and avoid artifacts.

\begin{figure}[tbp]
\centering\includegraphics[width=0.48\textwidth]{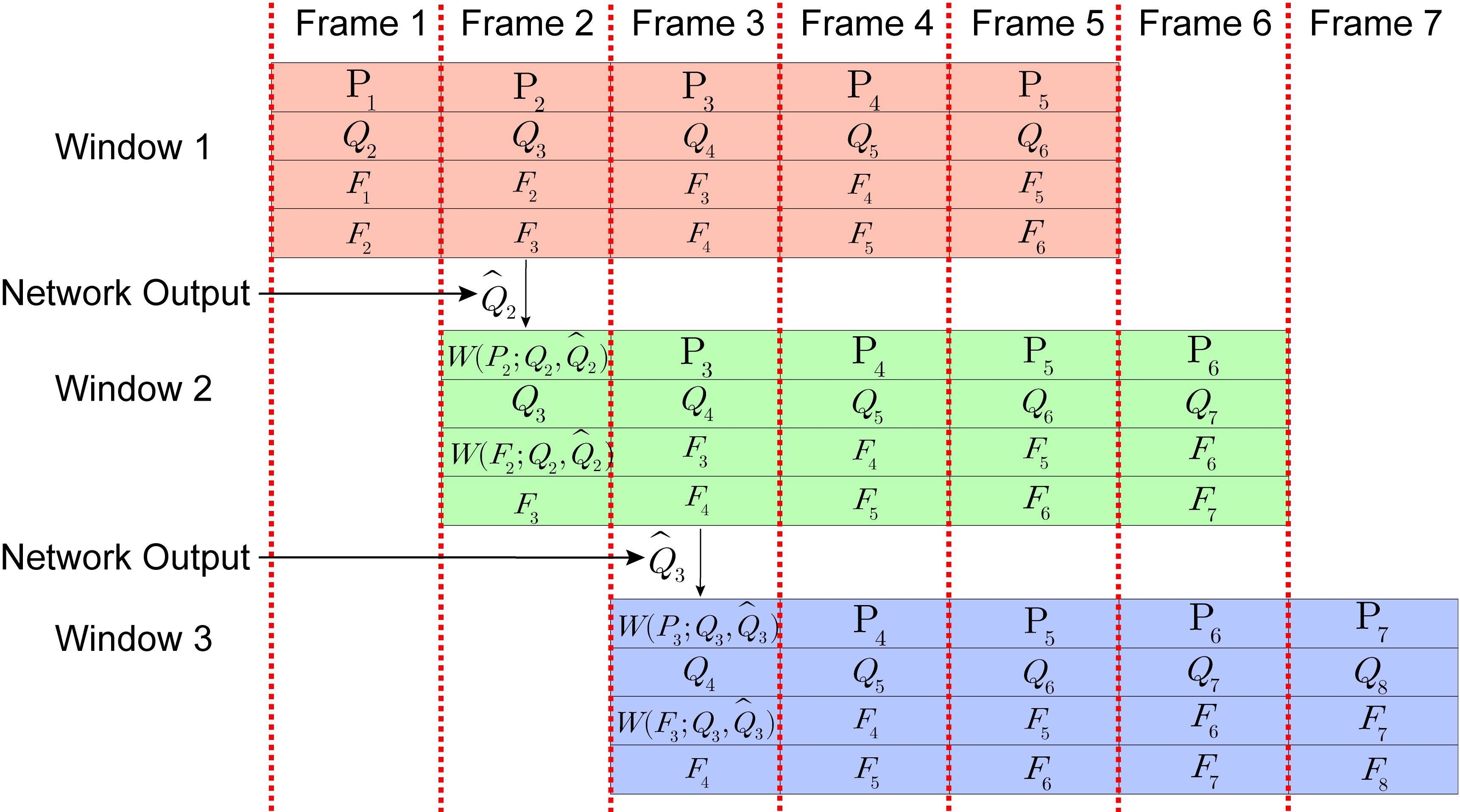}
\caption{\emph{The sliding window scheme of our method. The inputs of our network for each window are marked with the same color. For each window, the second frame is stabilized. The background feature points and the foreground face vertices are updated accordingly and become the next window's input.}}
\label{fig:sliding}
\end{figure}

\subsection{Sliding Window}\label{sec:slide}
Since the stabilization network only takes fixed length video segments, to apply to arbitrary length selfie videos, we apply a sliding window scheme.
In our experiment, we use a sliding window with length $T=5$.
We demonstrate our sliding window scheme in Fig.~\ref{fig:sliding}.
Each window is marked by the same color, which is the input to our network for the window.
Consider window 1 as an example.
The outputs of our stabilization network are the displacements of the warp nodes $\widehat{\mathbf{Q}}_2$, $\widehat{\mathbf{Q}}_3$ and $\widehat{\mathbf{Q}}_4$ as we discussed in the network structure.
At this point, if we warp all these frames in the current window and set the next window starting from frame 5, the result will be smooth within each window but not globally smooth.
Therefore, we only warp the second frame in the current window and shift one frame for the next window.
This scheme ensures temporal consistency between consecutive windows.
Specifically, in this example, we use the MLS warp function $W(\mathbf{v};\mathbf{Q}_2,\widehat{\mathbf{Q}}_2)$ to warp frame 2.
We then warp the feature points and face vertices using $W(\mathbf{P}_1;\mathbf{Q}_1,\widehat{\mathbf{Q}}_1)$ and $W(\mathbf{F}_1;\mathbf{Q}_1,\widehat{\mathbf{Q}}_1)$, since warping the frame leads to updated positions of the original feature points and face vertices.
The updated feature points and face vertices become a part of window 2, which is the next window starting at frame 2.

\section{Network Design}\label{sec:design}
In this section, we extend the discussion regarding the linear network design.
We first provide the complete list of parameters in the stabilization network in Sec.~\ref{subsec:parameter}.
We then discuss the necessity of using a network instead of formulating a linear optimization problem in Sec.~\ref{subsec:linear}.
In Sec.~\ref{subsec:optim}, we compare the performance of linear network and direct optimization of the loss function(Eq.~\ref{eqn:loss}).
In Sec.~\ref{subsec:nonlinear}, we compare the performance of linear network and non-linear network in terms of the quantitative metrics discussed in main paper Sec.~\ref{subsec:quant}.
Finally, we compare the performance using different number of filters in the network in Sec.~\ref{sec:filter}.

\subsection{Network Parameters}\label{subsec:parameter}
Table \ref{tab:network} lists the network parameters in main paper Fig.~\ref{fig:network}.
The number of filters in each layer is multiple of a base number $C$, which will be discussed in Sec.~\ref{sec:filter} in this supplementary material.

\setlength\tabcolsep{1pt}
\begin{table}[t]
\caption{Network parameters in main paper Fig.~\ref{fig:network}}
\label{tab:network}
\footnotesize
\begin{tabular}{p{0.035\textwidth}<{\centering}|p{0.07\textwidth}<{\centering}|p{0.073\textwidth}<{\centering}|p{0.073\textwidth}<{\centering}|p{0.045\textwidth}<{\centering}|p{0.045\textwidth}<{\centering}|p{0.05\textwidth}<{\centering}|p{0.05\textwidth}<{\centering}}
\hline
Layer id & Layer Type      & Input Size & Output Size & Kernel Size & Stride & Dilation & Padding \\ \hline
1     & Conv1d          & 4(T-1)x512 & Cx512       & 3           & 1      & 1        & 1       \\ \hline
2     & Conv1d          & Cx512      & 2Cx256      & 4           & 2      & 1        & 1       \\ \hline
3     & Conv1d          & 2Cx256     & 2Cx256      & 3           & 1      & 1        & 1       \\ \hline
4     & Conv1d          & 2Cx256     & 4Cx128      & 4           & 2      & 1        & 1       \\ \hline
5     & Conv1d          & 4Cx128     & 4Cx128      & 3           & 1      & 1        & 1       \\ \hline
6     & Conv1d          & 4Cx128     & 4Cx128      & 3           & 1      & 1        & 1       \\ \hline
7     & Conv1d          & 4Cx128     & 8Cx64       & 4           & 2      & 1        & 1       \\ \hline
8     & Conv1d          & 8Cx64      & 8Cx64       & 3           & 1      & 1        & 1       \\ \hline
9     & Conv1d          & 8Cx64      & 8Cx64       & 3           & 1      & 2        & 2       \\ \hline
10    & Conv1d          & 8Cx64      & 8Cx64       & 3           & 1      & 2        & 2       \\ \hline
11    & ConvT1d & 32Cx64     & 8Cx128      & 4           & 2      & 2        & 2       \\ \hline
12    & Conv1d          & 8Cx128     & 8Cx128      & 3           & 1      & 1        & 1       \\ \hline
13    & Conv1d          & 8Cx128     & 8Cx128      & 3           & 1      & 1        & 1       \\ \hline
14    & ConvT1d & 16Cx128    & 4Cx256      & 4           & 2      & 1        & 1       \\ \hline
15    & Conv1d          & 4Cx256     & 4Cx256      & 3           & 1      & 1        & 1       \\ \hline
16    & ConvT1d & 8Cx256     & 2Cx512      & 4           & 2      & 1        & 1       \\ \hline
17    & Conv1d          & 2Cx512     & 2Cx512      & 3           & 1      & 1        & 1       \\ \hline
18    & Conv1d          & 2Cx512     & 2(T-2)x512  & 1           & 1      & 1        & 0      \\ \hline
\end{tabular}
\end{table}

\subsection{Necessity of the linear network}\label{subsec:linear}
In Eq.~(\ref{eqn:lb}) and Eq.~(\ref{eqn:lf}) in the main paper, we define the loss function directly on feature points detected in the image.
This requires linear relationship between the input and the output of the stabilization network, i.e. scaling of feature point coordinates should lead to the same scaling of the output displacement to compensate the motion.
Note that this linear relationship between input and output can be posed as a matrix-vector product, i.e., $\mathbf{n}=\mathbf{Am}$ where $\mathbf{A}\in \mathbb{R}^{1024(T-1)\times 4096(T-1)}$ is a large matrix that transforms concatenated and reshaped input feature points and face vertices $\mathbf{m}\in \mathbb{R}^{4096(T-1)\times 1}$ to reshaped warp node displacements $\mathbf{n}\in \mathbb{R}^{1024(T-1)\times 1}$.
The optimization problem equivalent to our network training can be defined as:
\begin{equation}\label{eqn:optim}
\min_{\mathbf{A}}{L(\mathbf{m},\mathbf{n})},
\end{equation}
where $L$ is the loss function defined in Eq.~\ref{eqn:loss} in the main paper.
Solving this problem directly is difficult and prohibitive in the video stabilization for the following reasons.
First, the matrix $\mathbf{A}$ is dense and the problem is highly under-determined.
Second, the loss function we defined involves non-linear moving least squares warping; the problem cannot be solved using a simple linear system solver as in the bundled camera paths~\cite{bundle}.
Finally, the problem has to be solved for each sliding window in the online video stabilization, making it impossible to achieve real-time performance.
On the other hand, the linear neural network has two advantages compared to posing the problem as an optimization.
First, the convolutional layers contain only small kernels; the concatenation of layers is equivalent to decomposing the dense matrix into a series of sparse matrices which is easier to solve through backpropagation and gradient descent.
Second, the network implicitly provides regularization by training on a large dataset; using a pretrained network avoids the overfitting problem in the optimization and also enables computational real-time performance.

\begin{table}[t]%
\caption{Linear Network vs.\ Direct Optimization}
\label{tab:optim}
\begin{minipage}{\columnwidth}
\begin{center}
\begin{tabular}{c|c|c|c|}
\cline{2-4}

\multicolumn{1}{c|}{\textbf{Methods}}   & Cropping & Distortion & Stability \\ \hline
\multicolumn{1}{|c|}{Direct Optimization}   &  0.91  &  0.93   &  0.40   \\ \hline
\multicolumn{1}{|c|}{Our Linear Network} &  0.88  &   0.97  &  0.60   \\ \hline
\end{tabular}

\end{center}
\end{minipage}
\end{table}%

\begin{table}[t]%
\caption{Quantitative results from different network designs. In this table, $C$ is the number of filters in the first layer of our network depicted in Fig.~\ref{fig:network}}
\label{tab:design}
\begin{minipage}{\columnwidth}
\begin{center}
\begin{tabular}{clll}
\multicolumn{3}{c}{}                                                                                                                   \\ \cline{2-4} 
\multicolumn{1}{c|}{\textbf{C=32}}  & \multicolumn{1}{c|}{Cropping} & \multicolumn{1}{c|}{Distortion} & \multicolumn{1}{c|}{Stability} \\ \hline
\multicolumn{1}{|c|}{No activation} & \multicolumn{1}{c|}{0.85}         & \multicolumn{1}{c|}{0.95}           & \multicolumn{1}{c|}{0.56}          \\ \hline
\multicolumn{1}{|c|}{Leaky ReLU}    & \multicolumn{1}{c|}{0.90}         & \multicolumn{1}{c|}{0.97}           & \multicolumn{1}{c|}{0.48}          \\ \hline
\multicolumn{1}{|c|}{Tanh}          & \multicolumn{1}{c|}{0.87}         & \multicolumn{1}{c|}{0.97}           & \multicolumn{1}{c|}{0.50}          \\ \hline
\multicolumn{4}{c}{}                                                                                                                   \\ \cline{2-4} 
\multicolumn{1}{c|}{\textbf{C=64}}  & \multicolumn{1}{c|}{Cropping} & \multicolumn{1}{c|}{Distortion} & \multicolumn{1}{c|}{Stability} \\ \hline
\multicolumn{1}{|c|}{No activation} & \multicolumn{1}{c|}{0.86}         & \multicolumn{1}{c|}{0.96}           & \multicolumn{1}{c|}{0.57}          \\ \hline
\multicolumn{1}{|c|}{Leaky ReLU}    & \multicolumn{1}{c|}{0.92}         & \multicolumn{1}{c|}{0.98}           & \multicolumn{1}{c|}{0.52}          \\ \hline
\multicolumn{1}{|c|}{Tanh}          & \multicolumn{1}{c|}{0.85}         & \multicolumn{1}{c|}{0.96}           & \multicolumn{1}{c|}{0.52}          \\ \hline
\multicolumn{4}{c}{}                                                                                                                   \\ \cline{2-4} 
\multicolumn{1}{c|}{\textbf{C=128}} & \multicolumn{1}{c|}{Cropping} & \multicolumn{1}{c|}{Distortion} & \multicolumn{1}{c|}{Stability} \\ \hline
\multicolumn{1}{|c|}{No activation} & \multicolumn{1}{c|}{0.88}         & \multicolumn{1}{c|}{0.97}           & \multicolumn{1}{c|}{0.60}          \\ \hline
\multicolumn{1}{|c|}{Leaky ReLU}    & \multicolumn{1}{c|}{0.91}         & \multicolumn{1}{c|}{0.97}           & \multicolumn{1}{c|}{0.57}          \\ \hline
\multicolumn{1}{|c|}{Tanh}          & \multicolumn{1}{c|}{0.89}         & \multicolumn{1}{c|}{0.96}           & \multicolumn{1}{c|}{0.52}          \\ \hline
\multicolumn{4}{c}{}                                                                                                                   \\
\end{tabular}
\end{center}
\end{minipage}
\end{table}%
\begin{figure*}[tbp]
\centering\includegraphics[width=\textwidth]{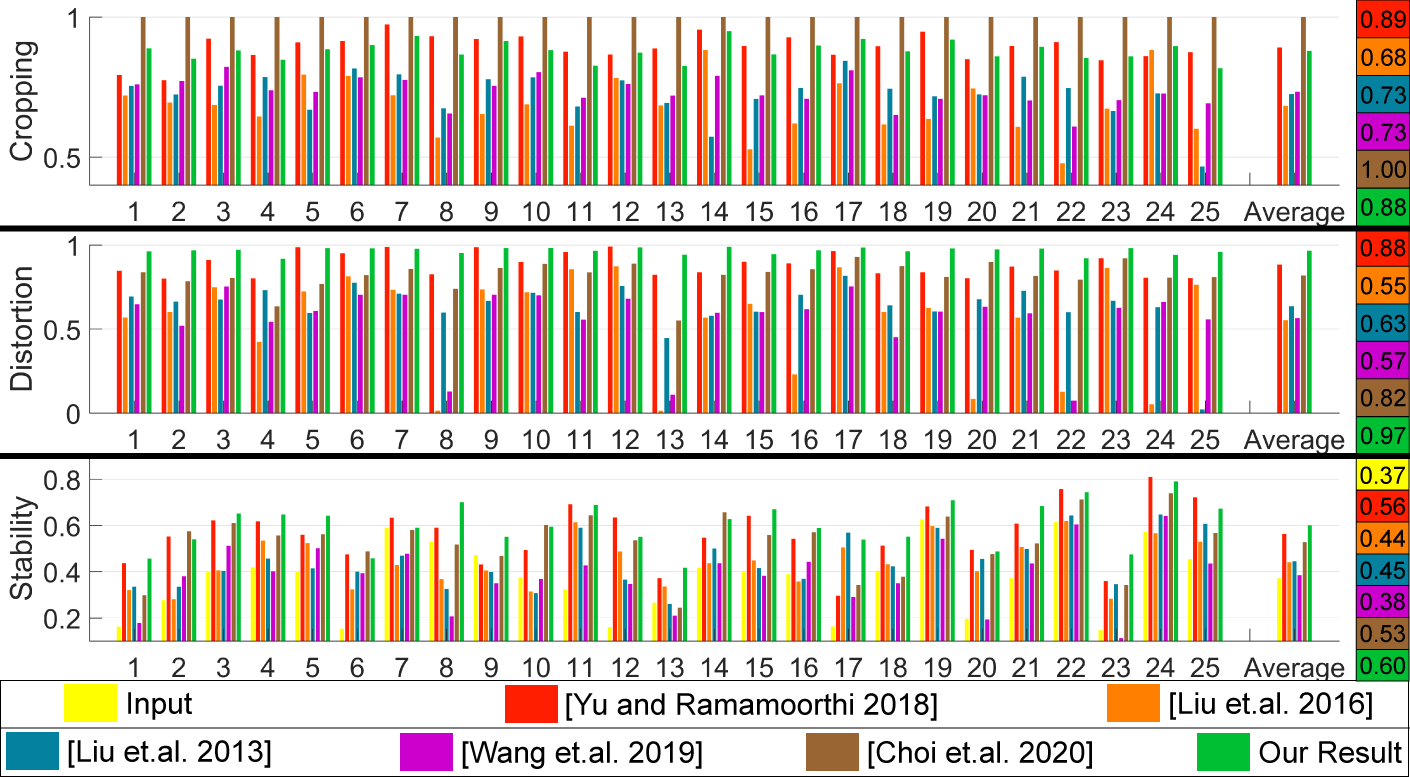}
\caption{\emph{Complete quantitative comparison of bundled camera paths~\cite{bundle}, selfie video stabilization~\cite{ourECCV}, MeshFlow~\cite{meshflow}, deep online video stabilization~\cite{wang}, deep iterative frame interpolation~\cite{choi} and our method. In these metrics, a larger value indicates a better result. }}
\label{fig:quant}
\vspace{-10pt}
\end{figure*}

\begin{figure}[tbp]
\centering\includegraphics[width=0.48\textwidth]{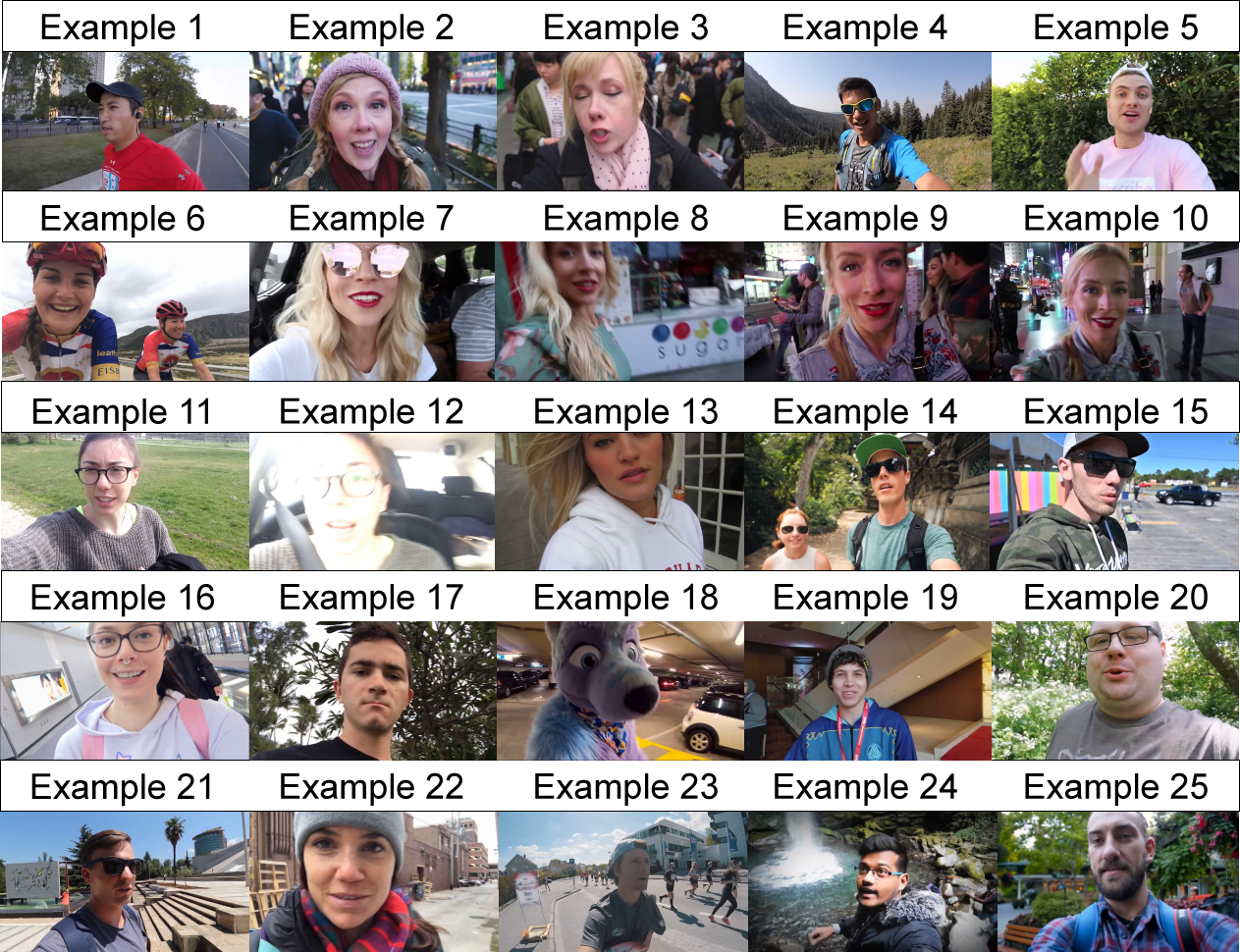}
\caption{\emph{The 25 selfie video examples used for testing, referred to in Fig.~\ref{fig:quant}}\newline}
\label{fig:example}
\vspace{-23pt}
\end{figure}
\subsection{Direct optimization}\label{subsec:optim}
Since our network is linear, an obvious question is whether we need a convolutional network at all.
A way to pose the stabilization process as an optimization problem is to directly solve for the warp node displacement $\widehat{\mathbf{Q}}_t-\mathbf{Q}_t$ to minimize the non-linear loss function $L$.
Note that the objective function $L$ is non-linear, so a simple least squares linear solver such as in bundled camera paths~\cite{bundle} cannot be used.  
We conduct an experiment in which we optimize our loss function Eq.~\ref{eqn:loss} in the main paper directly over the feature points (warp nodes) instead of network weights.
We optimize 1000 iterations using Adam optimizer with $lr=10^{-1}$, $\beta_1=0.9$ and $\beta_2=0.99$ for each 5-frame sliding window.
Note that although this formulation is tractable comparing to Sec.~\ref{subsec:linear}, the runtime of this optimization is prohibitive for pratical use since it requires an average of 20 seconds to stabilize each frame.
We show the quantitative comparison of this optimization result with the result generated by our linear network in Table~\ref{tab:optim}.
Although our network is linear, it performs significantly better than direct optimization.
This is expected; since the input feature points are sparsely distributed and the distribution varies frame from frame, blindly overfitting to the feature points in each sliding window will result in temporal inconsistency.
Our linear network provides implicit regularization for this process since it is trained over a variety of feature point distributions.
Therefore, this comparison proves that using the linear network is necessary and can produce significantly better results than optimization.

\begin{figure}[tbp]
\centering\includegraphics[width=0.48\textwidth]{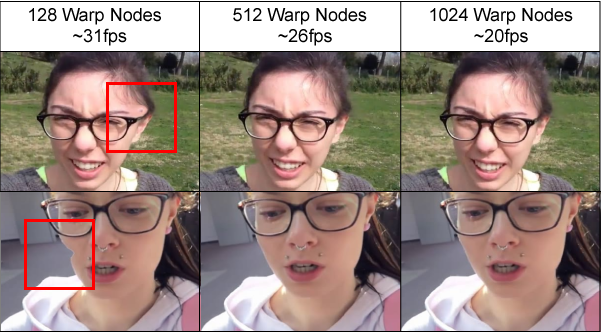}
\caption{\emph{The visual comparison and stabilization speed comparison of different number of warp nodes in our method. The artifacts are marked by the red box. }}
\label{fig:nodes}
\end{figure}

\subsection{Comparison with non-linear network}\label{subsec:nonlinear}
To justify our linear network design, we added different types of activation layers after each convolutional layer in our network and compare the result with our original network design.
To allow negative values in the network feature vectors, we select leaky ReLU and Tanh in our experiments. 
Table~\ref{tab:design} shows the averaged quantitative result over the examples in Fig.~\ref{fig:example} using the networks with leaky ReLU (with negative slope 0.2), tanh and no activation layers (our original design).
For the stability metric that is the most important, it can be observed that non-linear activation layers undermine the performance comparing to our original network design with the same base number of filters $C$.
The reason for this performance degradation is that the non-linear layers break the linear input/output relationship requirement.

\subsection{Number of filters}\label{sec:filter}
To show the effect of the number of filters used in each layer of the network, in Table~\ref{tab:design} we include the quantitative results with different numbers of filters in the input layer, i.e., $C=32, 64, 128$.
In general, the larger number of filters in the network, the better the results.
This conclusion also applies to the networks with non-linear activation layers, but the effect is more significant for the leaky ReLU activated network.
For the even more non-linear network with tanh layers, the performance saturates quickly with a greater number of filters $C$.
In this paper, we use $C=128$ in all the experiments.

\section{Additional Results}
In this section, we provide additional results to Sec.~\ref{sec:result} in the main paper.

\subsection{Complete Quantitative Result}\label{subsec:complete}
The complete list of video index and sample frames are shown in Fig.~\ref{fig:example}.
In Fig.~\ref{fig:quant}, we provide complete quantitative comparison with bundled camera paths~\cite{bundle}, selfie video stabilization~\cite{ourECCV}, MeshFlow~\cite{meshflow}, deep online video stabilization~\cite{wang}, deep iterative frame interpolation~\cite{choi}.
Note that we also modify the optimization based method bundled camera paths~\cite{bundle} by including the same mask detection procedure used in our pipeline, but it doesn't improve their result.

\begin{table}%
\caption{Ablation Study}
\label{tab:ablation}
\begin{minipage}{\columnwidth}
\begin{center}
\begin{tabular}{c|c|c|c|}
\cline{2-4}

\multicolumn{1}{c|}{\textbf{Ablation}}   & Cropping & Distortion & Stability \\ \hline
\multicolumn{1}{|c|}{No Foreground Detection} &  0.89  &   0.95  &  0.52   \\ \hline
\multicolumn{1}{|c|}{Full Pipeline}  &  0.88  &   0.97  &  0.60   \\ \hline
\end{tabular}

\end{center}
\end{minipage}
\end{table}%

\begin{table}%
\caption{Input Video Frame Size Comparison}
\label{tab:hd}
\begin{minipage}{\columnwidth}
\begin{center}
\begin{tabular}{c|c|c|c|}
\cline{2-4}

\multicolumn{1}{c|}{\textbf{Frame Sizes}}   & Cropping & Distortion & Stability \\ \hline
\multicolumn{1}{|c|}{HD ($1280\times720$)}   &  0.87  &   0.95  &  0.59   \\ \hline
\multicolumn{1}{|c|}{FHD ($1920\times1080$)} &  0.87  &   0.96  &  0.58   \\ \hline
\multicolumn{1}{|c|}{$832 \times 448$}  &  0.88  &   0.97  &  0.60   \\ \hline
\end{tabular}

\end{center}
\end{minipage}
\end{table}%

\subsection{The number of warp nodes (feature points)}\label{sec:warpnode}
The runtime performance of our method greatly depends on the number of warp nodes.
Note that we use the feature points as the warp nodes, therefore the number of warp nodes is equivalent to the number of feature points.
In the motion detection stage, tracking more feature points requires more processing time, leading to slower stabilization speed.
However, if the warp nodes are too sparse in the frame, the possibility of local distortion increases.
We provide the average per-frame stabilization time using 128, 512 and 1024 warp nodes and the corresponding warped frames in Fig.~\ref{fig:nodes}.
In Fig.~\ref{fig:nodes}, using 128 warp nodes results in distortion near the foreground/background bundaries.
This is because in the MLS warping, the warp nodes are implicitly constrained by each other.
Fewer constraints reduce the robustness of the warping.
An isolated warp node, if tracked mistakenly, introduces local distortion.
In our experiment, we select 512 warp nodes since it is a good balance between computational speed and warp quality.

\subsection{Ablation Study}\label{sec:ablation}
We performed an ablation study by removing the foreground mask detection stage in our pipeline.
This experiment means that we are essentially using all the feature points from both foreground and background, even if the foreground feature tracking is not reliable.
Table \ref{tab:ablation} shows the comparison with the full pipeline.
The stability score is significantly smaller than our full pipeline that separates the foreground and background.
However, note that even without foreground mask detection, we still outperform comparison optimization based methods ~\cite{meshflow,bundle}.
This also indicates that using the network is necessary for the video stabilization task.

\subsection{Video Frame Size}\label{sec:framesize}
The previously discussed results are tested with videos with frame size $832 \times 448$. 
Since our network only takes feature point/head vertices as the input, it is scalable with different frame sizes.
We tested our network with standard video resolutions (i.e., HD $1280\times720$ and Full HD $1920\times1080$) and compare the quantitative results with the $832 \times 448$ input, shown in Table~\ref{tab:hd}.
In these experiments, we resize the frame to $832 \times 448$ for faster feature detection and foreground/face detection. 
In the warping stage, we rescale the feature points and the output of our network.
Our network is able to handle higher resolution videos, and the result quality is similar to previously discussed results with frame size $832 \times 448$.




\end{document}